# Ignorance and the Expressiveness of Single- and Set-Valued Probability Models of Belief

Paul Snow
P.O. Box 6134
Concord, NH 03303-6134 USA
paulsnow@delphi.com

## Abstract

Over time, there have been refinements in the way that probability distributions are used for representing beliefs. Models which rely on single probability distributions depict a complete ordering among the propositions of interest, yet human beliefs are sometimes not completely ordered. Non-singleton sets of probability distributions can represent partially ordered beliefs. Convex sets are particularly convenient and expressive, but it is known that there are reasonable patterns of belief whose faithful representation require less restrictive sets. The present paper shows that prior ignorance about three or more exclusive alternatives and the emergence of partially ordered beliefs when evidence is obtained defy representation by any single set of distributions, but yield to a representation based on several sets. The partial order is shown to be a partial qualitative probability which shares some intuitively appealing attributes with probability distributions.

## 1. INTRODUCTION

Probability distributions have long been advocated as a useful foundation for the modeling of beliefs. The best known form of probabilistic belief representation consists of a single distribution. Such models bring with them a well-developed normative theory of behavior in the face of risk (Savage, 1972) which has had many adherents over the years.

Recently, some researchers have concluded that single distribution models are too restrictive. Beliefs may not always be completely ordered by the believer, even though a single probability distribution necessarily represents them as being so. Nevertheless, other attributes of probability distributions do seem like accurate portrayals of how beliefs behave with respect to Boolean combinations of the underlying events, and of how beliefs change in the face of evidence. Some of these desirable attributes are peculiar to probability distributions. So, to have the attributes, a belief representation must either use probability distributions or else use measures that agree with some probability distributions (Snow, 1992).

One way to get the desirable attributes of probabilities without the undesirable restrictiveness of a complete ordering is to model beliefs using non-singleton sets of probability distributions. It is often convenient to use convex sets of probability distributions, which arise as solutions to systems of simultaneous linear inequalities. Many natural language expressions of belief are easily translated into linear inequality constraints (Nilsson, 1986), e.g. "This event is at least as likely as that one." Linear constraint systems can be revised simply by Bayes' formula (Snow, 1991). Although there is a diversity of opinion about how set estimates might inform decision making, there are useful suggestions for decision rules in the literature (for a review, see Sterling and Morrell, 1991).

As versatile as convex sets are, there are reasonable belief patterns that convex sets fail to represent. For example, the set of posterior probabilities derived from a convex set of priors and a convex set of conditionals is generally not convex (White, 1986). Further, some important constraints are non-linear. Kyburg and Pittarelli (1992) discuss the non-convex sets which arise from the non-linear assumption of independence between events.

The present paper explores a circumstance where no single set of probability distributions, convex or otherwise, faithfully represents a reasonable pattern of belief, namely, ignorance being overcome by evidence when there are more than two alternatives. By *ignorance*, we mean that the believer is unwilling to assert any non-trivial prior ordering among the sentences of interest. By *being overcome by evidence*, we mean that the believer will assert some non-trivial orderings if the contrast between the conditional probabilities for the evidence given the sentences is sufficiently impressive.

A probabilistic solution to the representation of ignorance being overcome by evidence is presented. Although the model is more complex than a single set of probability distributions, the orderings that arise have much in common with single posterior probability distributions, and inference about the orderings is computationally inexpensive.

## 2. NOTATION AND ASSUMPTIONS ABOUT IGNORANCE

In this paper, we shall use the notation

$$S >e> T$$

to denote the condition that the believer asserts that sentence $S$ is, with a warrant satisfactory to the believer, at least as belief-worthy as sentence $T$ in light of evidence $e$. If evidence $e$ does not lead the believer to assert an ordering of sentences $S$ and $T$, then we write

$$S ?e? T$$

The condition of having no relevant evidence is indicated by the particle *nil*, as in

$$S ?nil? T$$

which expression denotes that there is no ordering between some sentences $S$ and $T$ in the absence of evidence.

We shall assume that the sentences of interest belong to a partitioned domain, which is defined as follows:

**Definition.** A *partitioned domain* is a set comprising:

(i) the always-true sentence, denoted **true**

(ii) the always false sentence, denoted **false**

(iii) two or more mutually exclusive sentences, called *atoms*

(iv) well-formed expressions involving atoms, **or**, and parentheses, called *simple disjunctions*

(v) well-formed expressions involving simple disjunctions, **true, false, or, not,** and parentheses

We shall assume throughout that the atoms in the domain are collectively exhaustive, that is, one of the atoms is true. This additional assumption places little epistemological burden on the believer (at worst, it means that one of the atoms is "none of the other atoms are true"), and has the convenient effect that every sentence in the domain has an equivalent simple disjunction. Finally, although infinite domains are useful in such applications as statistical hypothesis testing, we shall assume throughout this paper that the number of atoms in the domain is finite.

Our first assumptions about ignorance, and the conquest of ignorance by evidence express the following ideas. If no evidence has yet been observed, and the question of relative belief-worthiness is not answerable on logical grounds, then there is no satisfactory warrant to order one sentence ahead of another. Even after evidence has been observed, the question may remain open. Once a commitment to an ordering is made, then other commitments may be inferred by conditional probability considerations. A belief-ordering consistency principle discussed by Sugeno (unpublished dissertation, cited in Prade, 1985) obtains regardless of the presence or absence of evidence. The formal assumptions are:

**A1.** (Lack of explicit non-trivial prior orderings) For any sentences S and T,

$$S >nil> T \text{ implies that } T \text{ implies } S.$$

**A2.** (Lack of implicit non-trivial prior orderings) Values for conditional probabilities and orderings among them are neither known nor assumed if those values or orderings imply non-trivial constraints on the prior probabilities.

**A3.** (Consistency) For all evidence e, including *nil*, and any sentences S and T,

$$\text{if T implies S, then } S >e> T.$$

**A4.** (Impartiality) If $S >e> T$, and S' and T' are sentences, and S is exclusive of T, then

if S' is exclusive of T and $p(e \mid S') >= p(e \mid S)$, then $S' >e> T$, and

if S is exclusive of T' and $p(e \mid T) >= p(e \mid T')$, then $S >e> T'$.

**A5.** (Recovery from ignorance about atoms) For exclusive atoms s and t, and non-nil evidence e, a necessary condition for $s >e> t$ is that $p(e \mid s) >= p(e \mid t)$, and if $p(e \mid s) > 0$, then the inequality is strict. If $p(e \mid s) > 0$, then $p(e \mid t) = 0$ is not a necessary condition for $s >e> t$.

**A6.** (Dominance) For any sentences S, T, U and U' where (S **and** U) and (T **and** U) are both false and U' implies U, and for all evidence e, including *nil*,

if $( S \text{ or } U' ) >e> ( T \text{ or } U )$, then $S >e> T$, and

if $( S >e> T )$, then $( S \text{ or } U ) >e> ( T \text{ or } U' )$.

## 3. COMMENTARY ON THE ASSUMPTIONS

Assumption A1 explains one circumstance where we decline to assert any ordering: when there is no evidence, and the one sentence doesn't imply the other. A2 restricts the scope of the assumptions to problems whose givens rule out no prior probability distribution over the atoms. The conditions in assumption A2 reflect the easily-shown fact that a disjunctive conditional like $p(e \mid S)$ is a convex combination of the conditionals for the atoms in $S$, with weights proportional to the prior probabilities of the atoms.

Assumption A3 says that we always assert an ordering when one sentence implies another. This is Sugeno's consistency requirement.

Assumption A4 is about non-trivial orderings being "evidence driven" and about a different kind of consistency in the interpretation of evidence. Suppose $S$ comes to be ordered ahead of $T$ on account of evidence $e$, which is to say, not simply on the grounds of logical implication (since $S$ and $T$ are exclusive). If the evidence is even more favorable to $S'$ than $S$, or less favorable to $T'$ than $T$, then $S' >e> T$ or $S >e> T'$, provided that $S'$ and $T$ or $S$ and $T'$ have the same simple, exclusive relationship to one another that $S$ and $T$ do.

Assumption A5 establishes some conditions on how evidence can elicit belief in an ordering of atoms. The assumption requires that there actually be some advantage in the evidence (there are difficulties in allowing equal evidence strength to elicit ordering, to be discussed later in section 10), and also that so long as $s$ is not ruled out by the evidence, then it could be that $s >e> t$ without the evidence ruling $t$ out. The actual decision mechanism to assert that $s >e> t$ is deliberately left open. In principle, the decision would be based upon on the contrast between $p(e|s)$ and $p(e|t)$, perhaps founded on a standard of "interocular trauma" (in the memorable phrase of von Winterfeld and Edwards, 1986; the contrast "hits the believer between the eyes"). One specific decision rule will be developed for a particular model of the assumptions to be introduced shortly.

Assumption A6 can be viewed as a specialization of dominance notions, such as Savage's (1972) "Sure Thing Principle," to the inference, rather than decision, context. The assumption says that when comparing disjunctions, adding disjuncts to the favored side or removing disjuncts from the disfavored side does not disrupt the ordering. Nor is the ordering disrupted by adding or removing identical disjuncts from each side simultaneously.

## 4. ENSEMBLE OF SETS REPRESENTATION

One approach to modeling ignorance might be to say that the prior probability distribution over the atoms could be any probability distribution whatsoever. That is, we might use a set-valued prior estimate, the set of all probability distributions. As is typical in such estimates, $S$ would be asserted to be no less likely than $T$ when and only when $p(S) >= p(T)$ for all probability distributions over the atoms, that is, when all distributions in the set agree on the ordering in question. Call this the *unanimous agreement rule* for representing an ordering by a set of probability distributions.

Although that would surely represent a convincing degree of ignorance, it would also be **invincible** ignorance - no evidence short of the revelation of the truth about atoms for certain would ever overcome it. Except in cases where $T$ implies $S$ given the evidence (i.e., where there is no atom $t$ in $T$ that is not also in $S$, unless $p(e|t) = 0$), we would never assert $S >e> T$ in the sense that all our priors revise to a distribution where $p(S|e) >= p(T|e)$.

The least we could assume to enable vincible ignorance under the unanimous agreement rule is that there is some atom $s$ whose prior probability is positive. We needn't go so far as to assume that **all** atoms have a positive prior probability in order to decide an ordering question about any particular $S$ and $T$, but we do want to be able answer such questions for all $S$ and $T$ in the domain.

An ensemble of convex sets provides a way to represent ignorance for finite domains where evidence bears upon the atoms, and to make that ignorance vincible for all pairs of sentences on minimal assumptions for any particular pair. An ensemble contains sets of probability distributions. For every atom $s$, the ensemble contains the set of probability distributions which satisfy the constraints

$p(s) >= \pi$

$p(t) >= 0$     for all atoms $t$ besides $s$

$\Sigma p(\ ) = 1$     summation over all atoms

where $\pi$ is a constant which doesn't depend on $s$, and $\pi$ is both strictly greater than zero and strictly less than one-half.

If there are $N$ atoms in the domain, then there are $N$ vertices for each set in the ensemble. In the set for atom $s$, one vertex has unity for $p(s)$, and zeros for all other atoms' probabilities. All the other vertices have $p(s) = \pi$, and for one atom $t$ at each vertex, $p(t) = 1 - \pi$. All other atoms at such a vertex have zero probability.

The small number of vertices and their simple form make the application of Levi's (1980) procedure for the revision of convex sets (i.e., apply Bayes' formula to each vertex) especially easy. At each of the $N$-1 vertices with two non-zero elements, the posterior probabilities will be in the ratio

$$p(s|e)/p(t|e) = \pi/(1-\pi) * p(e|s)/p(e|t)$$

The minimal assumption for deciding the ordering between a pair of sentences is expressed in the following criterion. An ordering between sentences is asserted in this representation just in case that the ordering holds **for all the probabilities in at least one of the sets in the ensemble**. Of course, since all the sets are convex, this criterion is equivalent to requiring the ordering to hold for all the vertices in at least one set. $S\ ?e?\ T$ is asserted just in case that neither $S >e> T$ nor $T >e> S$ is asserted. We shall, of course, take $p(\text{true}) = 1$ and $p(\text{false}) = 0$ in all sets, and only sentences which are true or false given the evidence or *a priori* have a one or zero probability everywhere in any set.

## 5. DECIDING ORDERINGS IN AN ENSEMBLE OF SETS

To find out whether there is any set in an ensemble where $p(S|e) >= p(T|e)$ everywhere in the set can be accomplished by the following rules, which can be considered an implicit representation of the ensemble for computational purposes. There is no need for any explicit representation of the ensemble, and the only memory requirement is for the atoms themselves and the conditionals of the evidence.

**Theorem 1**. For simple disjunctions S and T, the following is an effective procedure to decide whether $S >e> T$:

(1) Eliminate atoms common to S and T, to create S' and T', the disjunction of atoms peculiar to S and T, respectively (if any)

(2) If S' and T' are both empty, or if evidence is not nil and there is no atom u in S' nor in T' where $p(e|u) > 0$, then $S >e> T$.

(3) If S' alone is empty and either evidence is nil or there is an atom t in T' where $p(e|t) > 0$, then **not** $S >e> T$.

(4) If T' alone is empty and either evidence is nil or there is an atom s in S' where $p(e|s) > 0$, then $S >e> T$.

(5) Otherwise, let s be an atom in S' where $p(e|s)$ is greatest among the atoms in S', and t be an atom in T' where $p(e|t)$ is greatest among the atoms in T'. If $p(e|s)/p(e|t) >= (1-\pi)/\pi$, then $S >e> T$, otherwise **not** $S >e> T$.

**Proof**. If the evidence is nil, then it is easy to show that $S >nil> T$ if and only if T implies S, and the requisite ordering is displayed by every probability distribution in the ensemble. Rule (1), and either (2) or (3), will apply.

Suppose, then, that the evidence is not nil. If T implies S given the evidence (that is, there is no atom t in T that is not also in S, except where $p(e|t) = 0$), then every probability distribution, and so every set in the ensemble, displays the ordering sought. Rule (1), and either (2) or (3) will apply. If S implies T given the evidence, and not the converse, then no set in the ensemble displays the ordering sought. Rules (1) and (4) will apply.

Suppose that neither S nor T imply the other given the evidence. The search for a set where the criterion holds can be restricted to the sets for atoms s in S.

The criterion for $S >e> T$ will fail to hold in any set for an atom s which is a disjunct in both non-equivalent disjunctions S and T. For S and T to be non-equivalent, there must be a state t in T that is not in S, and where

$p(e|t)$ is not zero. Therefore, in the set for a common atom s, there is a vertex where

$$p(S|e)/p(T|e) = \pi p(e|s) / [\pi p(e|s) + (1-\pi)p(e|t)]$$

which is strictly less than one. Rule (1) therefore discards no potential solution.

In applying the criterion for $S >e> T$, the search can be restricted to the set corresponding to the atom s in S' with the highest conditional evidence probability. If there is more than one such atom, then any one will do by the symmetry of the ensemble.

Within the set for that atom, attention can be restricted to the one vertex where the prior $p(t)$ is positive for the atom t in T' with the greatest conditional probability given the evidence e. At the vertex for t, if

$$p(e|s)/p(e|t) >= (1-\pi)/\pi$$

then the posteriors at this vertex are such that

$$p(s|e) >= p(t|e)$$

which means at this vertex that $p(S|e) >= p(T|e)$, since all the other priors are zero. If the order holds at the vertex for t, the vertex most favorable to T, then it holds at all the others for the set, and if it doesn't, then it doesn't hold everywhere in the set. Similarly, if the order doesn't hold everywhere in the most favorable set for S, then it doesn't hold everywhere in any other set. Rules (1) and (5) apply.   //

## 6. THE ENSEMBLE OF SETS FORMALISM IS A MODEL OF THE ASSUMPTIONS

**Theorem 2**. The ensemble of sets formalism is consistent with assumptions A1-A6.

**Proof**. (A1) The restriction on assertions from nil evidence is easily verified to hold within each set.

(A2) The assumption does not constrain the formalism, but rather the scope of problems to which the formalism may be applied. The restriction that all evidence bears directly on the atoms ensures that this constraint on the scope of application is respected.

(A3) Sugeno's consistency is an easily verified property of all probability distributions, and hence holds within each set.

(A4) If $S >e> T$ holds, and non-empty T excludes S, then the evidence is not nil, and there is some atom s in S and not in T, and $p(e|s)/p(e|t) >= (1-\pi)/\pi$. If $p(e|S') >= p(e|S)$, then it must be that for every atom in S' that $p(e|\text{atom in S'}) >= p(e|s)$ by A2 and the restriction that all evidence bears directly on the atoms. Since S' excludes T, there is some atom s' in S' that is not in T, and so $p(e|s')/p(e|t) \geq (1-\pi)/\pi$, and so the ordering holds in the set for every such s'. The argument for T' is similar.

(A5) If $p(e|s) < p(e|t)$, then it is easy to confirm that there is no set in the ensemble where $p(s|e) \geq p(t|e)$ at all the vertices. If $p(e|s) = p(e|t)$, then because $\pi$ is strictly less than one-half, $p(s|e)$ will be less than $p(t|e)$ at some vertex in each set unless $p(e|s) = 0$. Because $\pi$ is strictly positive, $p(s|e) \geq p(t|e)$ at every vertex in the set for atom s for some positive value of $p(e|t)$ whenever $p(e|s) > 0$.

(A6) Dominance of this kind is a property of every probability distribution. To show that S or U' >e> T or U implies S >e> T: there must be some set in the ensemble where $p(S \text{ or } U' | e) \geq p(T \text{ or } U | e)$ holds for every distribution in the set. So, $p(S|e) \geq p(T|e)$ everywhere in that set as well. The argument for the other condition is similar. //

## 7. SOME OTHER PROBABILISTIC FORMALISMS WHICH ARE NOT MODELS OF THE ASSUMPTIONS

The ensemble of sets formalism is more complex than using either a single probability distribution or else a non-singleton set of probabilities as an uninformative prior to represent initial ignorance which is overcome by subsequent evidence through Bayes' Formula. Both of these simpler structures (under ordinary ways to interpret how orderings are expressed) fail to conform to one or more of the assumptions A1-A6 in partitioned domains with three or more atoms.

In the case of a single distribution prior, assuming that S >nil> T just in case that $p(S) \geq p(T)$, then assumption A1 is violated. Probabilities in a single distribution are completely ordered. So, even before any evidence is seen, for all sentences S and T, S >nil> T, or T >nil> S, or both - not just for sentences where one implies the other.

The same point is often argued on intuitive grounds for the Principle of Insufficient Reason when there are three or more atoms in the partitioned domain. If $p(s) = p(t) = p(u)$, then *a priori* $p(s \text{ or } t) > p(u)$. Assumption A1, of course, is violated even in dichotomous domains, since if $p(s) = p(\text{not } s)$, then s >nil> not s and not s >nil> s, even though neither atom implies the other.

Any single distribution (not just the PIR distribution) in a partitioned domain with three or more distinct atoms (s, t, u, ...) need not conform to assumption A4 (Impartiality). Suppose the evidence is such that $p(e|s) = p(e|t) > 0$. Standard results show that $p(e|s \text{ or } t)$ is also equal to $p(e|s)$ regardless of the prior probabilities of s and t. Assuming that none of the priors is zero (required by assumption A5), then there are values of $p(e|u)$ and $p(e|s)$ such that

$$p(u|e) > p(s|e)$$

but

$$p(u|e) < p(s \text{ or } t | e)$$

even though the evidence for s or t is no stronger than that for s alone. The conclusion that s or t was no less likely than u would not be supported by the evidence in such a case.

The situation for set-valued priors is better. Consider the set of distributions that solve the N (= number of atoms in the domain) simultaneous linear constraints

$$p(s) \geq c \quad \text{for all atoms s}, 0 < c < 1/(2N-2) \quad [1]$$

and total probability, where S >e> T just in case $p(S|e) \geq p(T|e)$ in every probability distribution in the solution set. Bayesian revision of [1] can be accomplished by applying Bayes formula to the N vertices of the solution set, as explained earlier in connection with the ensemble of sets formalism. Each vertex has exactly two non-zero components.

Here there is no trivial ordering is asserted on nil evidence. It is easy to verify that for any sentences S and T unrelated by implication, there is a distribution where $p(S) > p(T)$ and a distribution where $p(S) < p(T)$, so no ordering is asserted under the "unanimous agreement" rule.

In dichotomous domains, the single convex set [1] and the ensemble of sets where $\pi = c$ are easily shown the be equivalent. Any ordering asserted in the one is asserted in the other. Dichotomous domains are an important special case. For example, much of the plausible reasoning about mathematical propositions studied by Polya (1954) concerns dichotomous domains.

With several atoms in a domain, however, there are practical problems with the single set. Analysis of the vertices of [1] shows that the smallest Bayes Factor needed to assert s >e> not s for any atom cannot be less than N - 1. This is a nuisance: after all, in a domain with 100 atoms, many people would experience interocular trauma with a much smaller Bayes Factor than 99. The dependence of the evidentiary contrast required for ordering assertions upon the number of atoms would inhibit the application of system [1] to large domains.

In fact, all set-valued priors, including but not limited to system [1], can violate the assumptions for some evidence when there are three or more atoms in the domain. The argument is similar to that made earlier for a single-distribution prior in similar domains. Consider the distribution in the set most favorable to s compared to u, that is, where p( s ) / p( u ) is greatest. Assumption A5 requires that none of the priors be zero, so if p( e | s ) = p( e | t ) = p( e | s or t ) as before, then there exist p( e | u ) and p( e | s ) such that

p( u | e ) > p( s | e ) and p( u | e ) < p( s or t | e )

within this distribution. Since this is the distribution most favorable to *s* compared to *u*, *u* is asserted to be no less likely than *s* (if the ordering holds here, then it holds everywhere in the set), but it is not asserted that *u* is no less likely than *s or t* (since the ordering doesn't hold here, it doesn't hold everywhere), even though the evidence is no stronger for *s or t* than for *s* alone.

We now return to the ensemble of sets formalism, and show that the orderings that it permits, while not generally equivalent to any probability distribution, are nevertheless intuitively appealing from a probabilist perspective.

## 8. PARTIAL QUALITATIVE PROBABILITY

**Definition.** A *partial qualitative probability* is a partial order of the sentences in a domain, such that, for all evidence e, including nil, and any sentences S, T, and U:

(i) (boundedness) **true** >e> S and S >e> **false**

(ii) (transitivity) ( S >e> T ) **and** ( T >e> U ) implies S >e> U

(iii) (quasi-additivity) if S **and** U and T **and** U are both false, then

( S or U ) >e> ( T or U ) if and only if S >e> T.

This definition is designed to echo that of an ordinary qualitative probability (de Finetti, 1937). Qualitative probabilities abstract some of the ordering properties of ordinary probability distributions, and every probability distribution is also a qualitative probability. Partial qualitative probabilities as defined here differ from ordinary qualitative probabilities only in being partial, rather than complete, orderings. Partial qualitative probabilities can also be shown to display other intuitively appealing properties (to a probabilist, at least) beyond those used in the definition, such as complementarity:

S >e> T implies **not** ( T ) >e> **not** ( S )

## 9. THE ENSEMBLE OF SETS FORMALISM IS A PARTIAL QUALITATIVE PROBABILITY

**Lemma.** If A, B, C, and D are simple or empty (containing no atoms except those where p( e | atom ) = 0) disjunctions where there is no atom in common between A and B, nor any atom in common between C and D, then in the ensemble of sets formalism

A >e> B **and** C >e> D implies A or C >e> B or D

**Proof.** If ( B or D ) implies ( A or C ), then the required ordering holds. Suppose that is not the case. If B is empty or D is empty, then the lemma is trivial. If A is empty, then B is empty, and if C is empty, then D is empty. Suppose none of them are empty. For orderings to be asserted, evidence must be non-nil. Let a, b, c, and d be the atoms such that p( e | atom ) is greatest among atoms in A, B, C, and D respectively. WOLG, suppose that p( e | a ) >= p( e | c ). Since p( e | c ) / p( e | d ) >= ( 1 - $\pi$ ) / $\pi$, then p( e | a ) / p( e | d ) >= ( 1 - $\pi$ ) / $\pi$, and the inequality holds for a and every atom in D. Since $\pi$ < 1/2, atoms a and d have different p( e | atom )'s, and so they are distinct, and a is also distinct from every atom in D; a and the atoms of B are distinct by hypothesis, so a is not eliminated by step (1) of theorem 1, and the required ordering is asserted in the set for a. //

**Theorem 4.** The ensemble of sets formalism is a partial qualitative probability.

**Proof.** It can be shown that any ordering that satisfies assumptions A1-A6 is a partial qualitative probability. It is also easy to verify that the special case of the ensemble of sets formalism in particular is a partial qualitative probability.

**Boundedness**: is a property of all probability distributions, and so p( **true** ) >= p( S ) and p( S ) >= p( **false** ) everywhere in all sets in the ensemble, and so the required orderings are asserted.

**Quasi-additivity**: Quasi-additivity is a property of every probability distribution. To show that S or U >e> T or U implies S >e> T: there must be some set in the ensemble where p( S or U | e ) >= p( T or U | e ) holds for every distribution in the set. So, p( S | e ) >= p( T | e ) everywhere in that set as well. The argument for the converse is similar.

**Transitivity**: Define the following sets: S* = { S and not T and not U }, ST = { S and T and not U }, SU = { S and U and not T }, T* = { T and not S and not U }, TU = { T and U and not S }, and U = { U and not S and not T }. Some of these sets may be empty, but note that in the absence of implications, there must be at least one distinct atom on each side of the ">e>" operator in any asserted ordering.

By quasi-additivity, S >e> T if and only if

S* or SU >e> T* or TU    [2]

No atom is common to any two sets in [2]. Similarly, T >e> U if and only if

T* or ST >e> U* or SU    [3]

No atom is common to any two sets in [3].

To show S >e> U, we show S* or ST >e> U* or TU. Applying Lemma 2 to [2] and [3]:

S* or SU or T* or ST >e> T* or TU or U* or SU

which, by quasi-additivity, reduces to the desired expression.    //

## 10. A NOTE ON ASSUMPTION A5

In assumption A5, we required that p( e | s ) be strictly greater than p( e | t ) in order for s >e> t to hold. The requirement was echoed in the definition of the ensemble of sets formalism by the provision that $\pi$ should be strictly less than one-half. We now present an example where if A5 called for a weak inequality, the resulting ordering would fail to be a partial qualitative probability.

Suppose there are six distinct atoms in a partitioned domain, $i$ through $n$, and evidence conditionals

p( e | i ) = .6    p( e | k ) = .5    p( e | m ) = .4
p( e | j ) = .4    p( e | l ) = .6    p( e | n ) = .5

Consider the disjunctions S = i or j or m, T = k or l or m, and U = i or n or k, and suppose that ordering is asserted on equal conditionals. S >e> T, since by theorem 1 or the quasi-additivity property, we compare $i$ or $j$ (where $i$ has the higher conditional, .6) with $k$ or $l$ (where $l$ has the higher, also .6). Similarly T >e> U, since we compare $l$ or $m$ ($l$ has the higher, .6) with $i$ or $n$ ($i$ has the higher, again .6). If the ordering were a partial qualitative probability, then S >e> U, but this is not so. Comparing $j$ or $m$ (both have .4) with $n$ or $k$ (both have .5), we fail to assert S >e> U.

It can be shown that this feature is not peculiar to the ensemble of sets formalism, but is displayed by any ordering that obeys the weakened A5 and the other assumptions. Note also that the strict inequality requirement, which is implemented in the ensemble of sets by requiring that $\pi$ be strictly less than one-half, forecloses the possibility that for exclusive sentences $S$ and $T$, p( S | e ) >= p( T | e ) everywhere in some set in the ensemble, while p( T | e ) >= p( S | e ) everywhere in any set in the ensemble. The proof, which is a straightforward application of Theorem 1, especially rule (5), is omitted.

## 11. CONCLUSIONS

In real life, beliefs have great subtlety. Although probability distributions possess intuitively appealing properties that seem to capture some aspects of belief, single distributions, convex sets of distributions, and now general single sets of probability all fall short of doing justice to that subtlety.

The circumstance studied in this paper where single sets fall short, ignorance followed by the advent of partially ordered belief, is frequently encountered in important practical situations. A more general probabilistic structure, the ensemble of sets formalism, appears adequate to model beliefs in that circumstance. The orderings that evidence elicits within the formalism are intuitively appealing from a probabilistic perspective, and the computational effort required to decide ordering questions in partitioned domains is modest.